\documentclass[]{kling/kling}

\usepackage{microtype}
\usepackage{graphicx}
\usepackage{subcaption}
\usepackage{csquotes}
\usepackage{afterpage}
\usepackage{xcolor}

\usepackage{amsmath}
\usepackage{amssymb}
\usepackage{mathtools}
\usepackage{amsthm}
\usepackage{xspace}
\usepackage{colortbl}
\usepackage{multirow}
\usepackage{enumitem}
\usepackage{wrapfig}
\usepackage{nicefrac}
\usepackage[font=small,labelfont=bf,justification=justified]{caption}

\usepackage{booktabs}
\usepackage{multirow}
\usepackage{graphicx}
\usepackage{xcolor}
\usepackage{pifont} %

\usepackage{geometry}
\usepackage{tabularx}

\usepackage{amsmath}

\newcommand{\methodname}{Kling-MotionControl}
\newcommand{\myparagraph}[1]{\noindent\textbf{#1}}

\title{Kling-MotionControl Technical Report}
\headertext{Kling-MotionControl: Orchestrating Heterogeneous Motions for Adaptive Holistic Character Animation}

\author[]{Kling Team, Kuaishou Technology}

\abstract{
Character animation aims to generate lifelike videos by transferring motion dynamics from a driving video to a reference image. Recent strides in generative models have paved the way for high-fidelity character animation. In this work, we present \textbf{\methodname{}}, a unified DiT-based framework engineered specifically for robust, precise, and expressive holistic character animation. Leveraging a ``divide-and-conquer'' strategy within a cohesive system, the model orchestrates heterogeneous motion representations tailored to the distinct characteristics of body, face, and hands, effectively reconciling large-scale structural stability with fine-grained articulatory expressiveness. To ensure robust cross-identity generalization, we incorporate adaptive identity-agnostic learning, facilitating natural motion retargeting for diverse characters ranging from realistic humans to stylized cartoons. Simultaneously, we guarantee faithful appearance preservation through meticulous identity injection and fusion designs, further supported by a subject library mechanism that leverages comprehensive reference contexts. Furthermore, we enhance our proposed motion representations with 3D awareness, enabling precise alignment across diverse character orientations and flexible cinematic camera control via native text descriptions. To ensure practical utility, we implement an advanced acceleration framework utilizing multi-stage distillation, boosting inference speed by over 10$\times$. \methodname{} distinguishes itself through intelligent semantic motion understanding and precise text responsiveness, allowing for flexible control beyond visual inputs. Human preference evaluations demonstrate that \methodname{} delivers superior performance compared to leading commercial and open-source solutions, achieving exceptional fidelity in holistic motion control, open domain generalization, and visual quality and coherence. These results establish \methodname{} as a robust solution for high-quality, controllable, and lifelike character animation.
}

\metadata[Date]{March 3, 2026}
\metadata[Access]{\url{ https://app.klingai.com/global/video-motion-control/new}}

\begin{document}

\maketitle

\section{Introduction}
\vspace{-.5em}
\begin{figure}[t!]
    \centering
    \includegraphics[width=\linewidth]{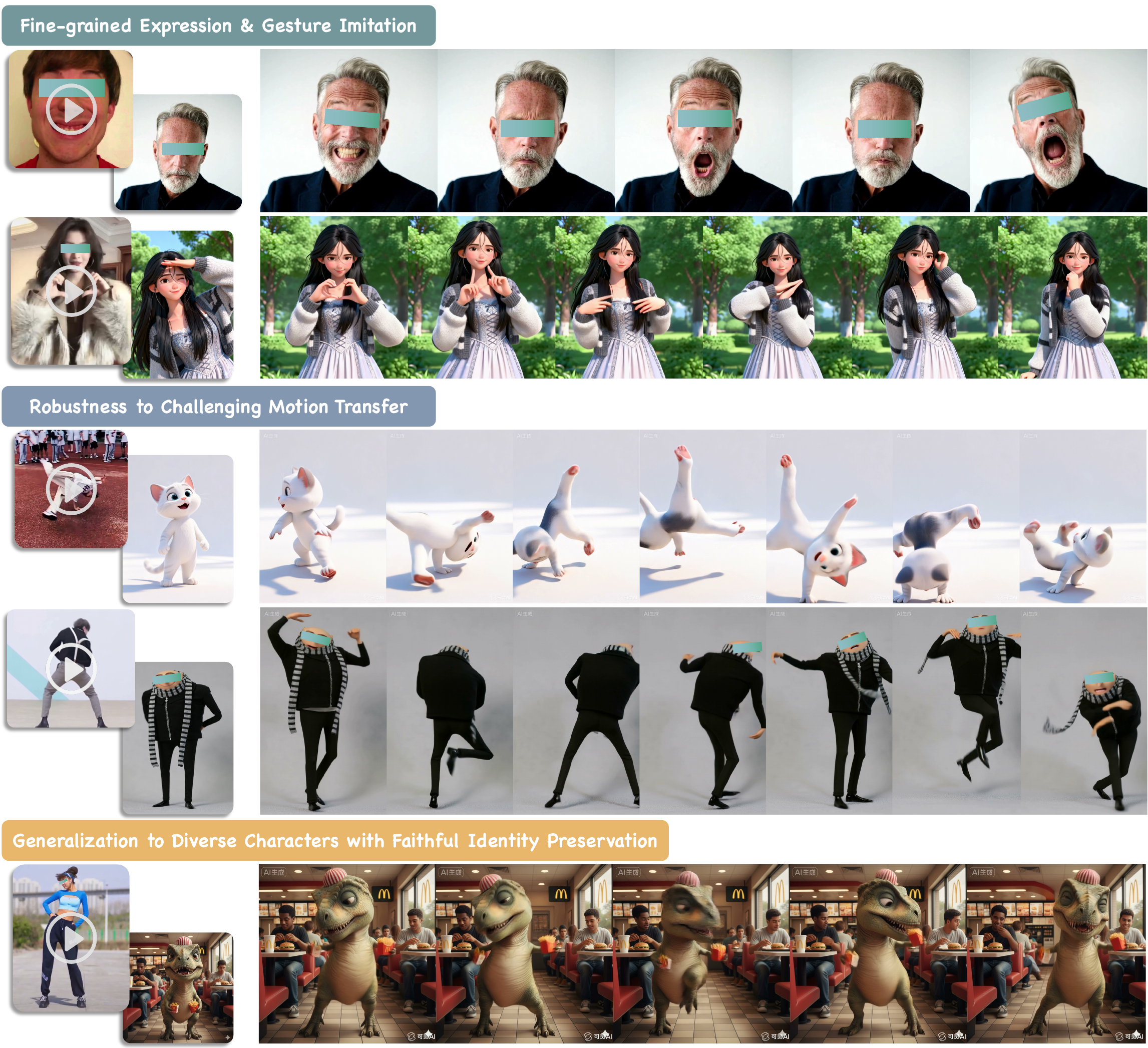}
    \caption{Given a reference image and a driving video, \textbf{\methodname{}} generates high-fidelity videos where the reference subject faithfully mimics multi-granular driving motions, encompassing body movements, facial expressions, and hand gestures.
    Our results demonstrate precise fine-grained control, robustness against rapid and complex dynamics, and natural cross-identity transfer with faithful identity preservation, generalizing seamlessly to diverse characters such as anime, cartoons, and stylized artworks.
    }
    \label{fig:teaser}
    \vspace{-1.2em}
\end{figure}

Character image animation aims to generate animated videos by transferring motion dynamics from a driving video to a reference image containing a distinct subject~\citep{cheng2025wananimate,wang2025unianimatedit,luo2025dreamactor}. This technology holds widespread potential for applications ranging from digital avatar creation to animation production and controllable video synthesis. The fundamental objective is to accurately model the motion dynamics from a driving sequence and seamlessly adapt them to a novel character, achieving precise motion control while faithfully preserving the visual appearance of the reference~\citep{animateanyone}.

Recent advances in large-scale video generative models~\citep{wan2025wan,kong2024hunyuanvideo,li2024sora}, particularly Diffusion Transformers (DiTs)~\citep{peebles2023scalable}, have substantially advanced the field of character animation from a single image. Nevertheless, early approaches predominantly focused on either facial reenactment~\citep{guo2024liveportrait,xie2024xportrait,qiu2025skyreels} or body motion control~\citep{animateanyone,zhu2024champ,zhang2024mimicmotion} in isolation. While recent endeavors, such as Dreamina~\citep{dreamina}, Runway Act-Two~\citep{runway_act_two}, and Wan-Animate~\citep{cheng2025wananimate}, have begun to explore holistic full-body animation, they often struggle to effectively reconcile visual quality with controllability across varying motion granularities. Specifically, these methods frequently exhibit limitations in balancing large-scale limb stability with fine-grained details (e.g., facial micro-expressions and finger articulation), and suffer from identity drift during cross-identity transfer, especially with diverse morphologies like anime or animals. Moreover, they often lose control over other visual attributes (e.g., background, camera moving) when prioritizing motion constraints. Furthermore, the prohibitive computational cost and limited inference efficiency remain critical bottlenecks impeding the practical deployment of these high-fidelity models.

In this work, we present \textbf{Kling-MotionControl}, a unified DiT-based framework designed for robust, precise, expressive, and efficient holistic character animation. Beyond serving as a robust motion transfer tool, it functions as an intelligent and controllable system capable of handling comprehensive animation scenarios, from full-body motion transfer to fine-grained facial reenactment, while maintaining exceptional identity fidelity and inference efficiency. \textbf{Kling-MotionControl} incorporates the following key technical advancements:

\begin{figure}[t!]
    \centering
    \captionsetup{justification=centering} 
    
    \includegraphics[width=\linewidth]{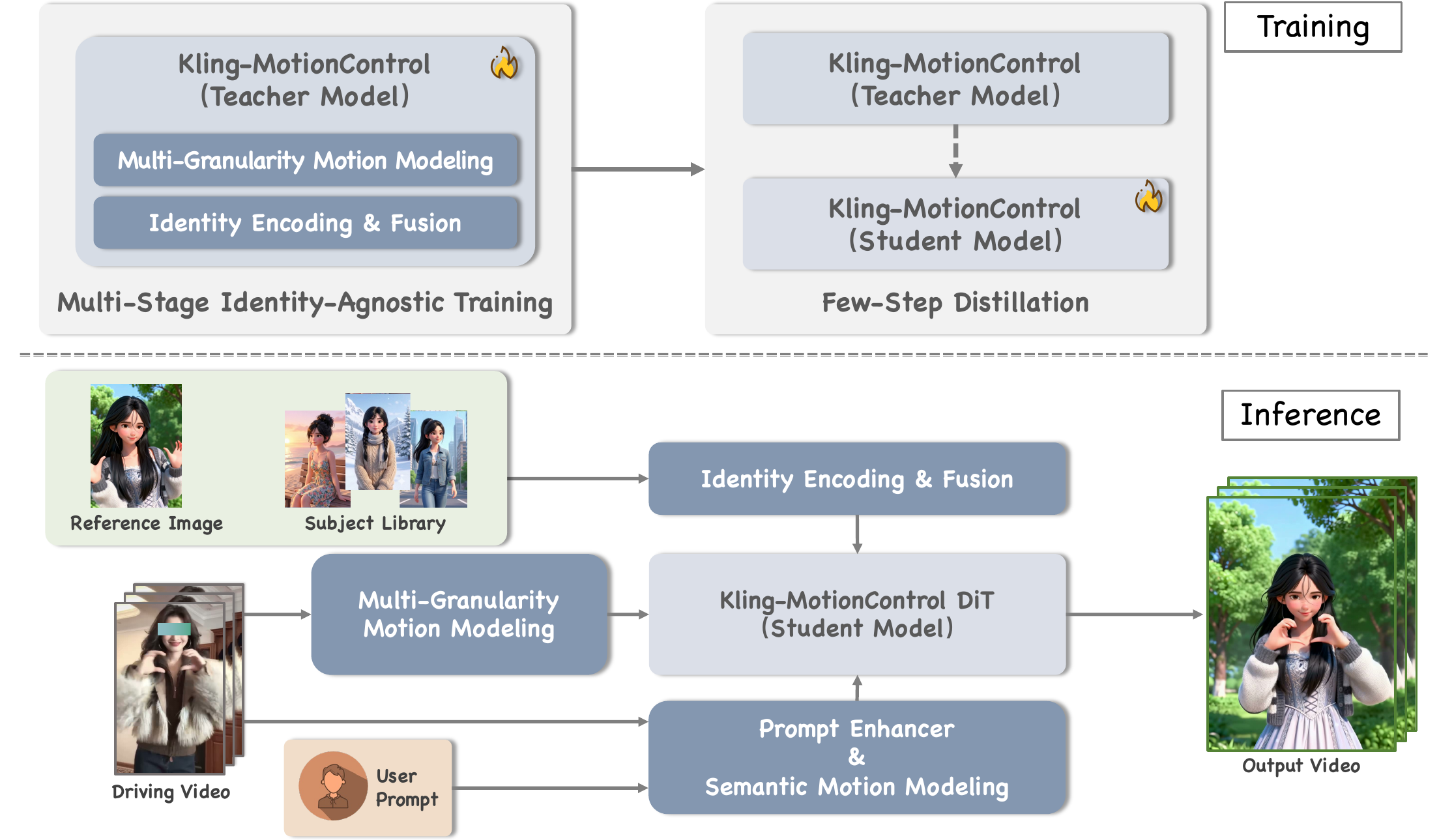}
    \caption{Overview of training and inference pipeline of our proposed \methodname{}.}
    \label{fig:teaser}
\end{figure}

\begin{itemize}
    \item \textbf{Unified Multi-Granularity Motion Orchestration.} \methodname{} is a unified framework that orchestrates \textbf{heterogeneous motion representations} tailored to the inherent characteristics of distinct motion granularities---specifically the body, face, and hands. Harmonized via a \textbf{progressive multi-stage training strategy}, this ``divide-and-conquer'' approach enables the cohesive modeling of both the structural stability required for large-scale body movements and the delicate expressiveness needed for facial micro-expressions and intricate finger interactions. This design facilitates seamless motion transfer across various scales, from close-up portraits to dynamic full-body scenes within a single unified model. Crucially, it effectively overcomes the robustness issues in large-amplitude movements and the lack of precision in fine-grained details often observed in previous works, thereby minimizing visual artifacts while ensuring coherent coordination among facial expressions, hand gestures, and body poses.

    \item \textbf{Adaptive Cross-Identity Motion Transfer.} To support robust generalization across diverse characters ranging from realistic humans to stylized cartoons and even animals, we introduce an \textbf{identity-agnostic motion learning} paradigm. This approach distills the essence of motion by decoupling dynamic patterns from the driving subject's physical attributes at the geometric level. Complementing this geometric abstraction, we further incorporate a \textbf{semantic motion modeling module} to capture the high-level intent of actions (e.g., ``facepalm'', ``clapping''). This ensures that the generated animation is not only geometrically aligned but also semantically faithful to the driving performance, effectively resolving ambiguities in complex interactions. Consequently, \methodname{} achieves natural motion retargeting across significant morphological disparities (e.g., adult-to-child, human-to-animal) without requiring manual calibration.

    \item \textbf{Faithful Identity Preservation with subject library Support.} \methodname{} achieves superior identity fidelity through a dedicated \textbf{identity encoding and fusion} mechanism. By meticulously extracting and integrating identity embeddings, \methodname{} effectively ensures the reference character's traits are strictly maintained during transfer. Furthermore, to enhance robustness in complex scenarios, \methodname{} supports a \textbf{subject library} mechanism. Unlike standard single-image approaches, this feature allows users to provide additional reference materials, such as multi-view images or video clips of the target character. This comprehensive context enables the model to construct a more robust identity representation, ensuring exceptional subject consistency and strictly preserving appearance details even during extreme poses or long-duration generation.

    \item \textbf{3D Awareness with Free-View Camera Control.} The multi-granular motion representations are further endowed with 3D perception capabilities through large-scale multi-view supervision. This paradigm enables \methodname{} to perceive the intrinsic 3D geometry and dynamics of the driving motion beyond simple 2D plane alignment, supporting the flexible specification of character orientations in the animated results. Furthermore, it empowers flexible cinematic camera control, allowing users to perform free-view rendering with dynamic camera trajectories (e.g., pans, zooms) controlled directly via native text descriptions, while maintaining strict geometric consistency and structural integrity.

    \item \textbf{Intelligent Text Responsiveness.} \methodname{} utilizes an intelligent \textbf{Prompt Enhancer (PE)} module to bridge the gap between motion control and textual guidance. This enables the model to maintain precise motion adherence while remaining highly responsive to user text prompts. Users can flexibly manipulate scene elements, clothing styles, and camera movements via text, ensuring a high degree of creative controllability beyond the reference image. 

    \item \textbf{High-Efficiency Inference Acceleration.} Addressing the high computational cost of video generation, we implement an advanced acceleration framework. We first introduce an efficient \textbf{dual-branch sampling strategy} for the teacher model to handle multi-conditional Classifier-Free Guidance (CFG) without the computational burden of multiple inference branches. To further compress the generation process, we optimize a multi-stage \textbf{distillation} strategy that substantially reduces the Number of Function Evaluations (NFE), yielding a high-quality few-step student model. Moreover, by merging conditional gradients into the student model, we effectively bypass the sampling overhead typically associated with CFG. These comprehensive optimizations achieve an end-to-end acceleration exceeding 10$\times$ while preserving model performance, significantly lowering deployment costs.

    \item \textbf{Comprehensive Data Curation Framework.} To empower the model's robust capabilities, we present a holistic data framework integrating a dedicated curation pipeline and a scalable annotation infrastructure. We have collected a massive dataset encompassing a wide spectrum of character types and diverse motion dynamics. To ensure data excellence, we implement a rigorous filtering process based on multi-dimensional criteria, encompassing key indicators such as overall video quality scores, motion dynamics (e.g., amplitude and fluency), and subject consistency. Furthermore, this dataset is supplemented by high-quality rendering data and footage captured via high-speed cameras to support the optimization of rapid and complex movements. Finally, our fine-grained annotation system covers detailed attributes including specific actions, micro-expressions, human-object interactions, and camera moves, providing rich guidance to facilitate robust model training for professional-grade animation generation.
\end{itemize}

We envision \textbf{Kling-MotionControl} serving as a vital productivity tool, designed to enhance both professional animation workflows and daily creative applications by delivering cinematic-quality motion control with unprecedented efficiency and flexibility.

\section{Evaluation}
\subsection{Evaluation Settings}
To ensure a comprehensive evaluation, we constructed a dedicated benchmark consisting of 150 high-quality test cases, each featuring a reference image paired with a driving video from a distinct subject.
We adopt a human preference-based subjective evaluation protocol as our assessment standard, aiming to accurately capture user-perceived perceptual quality and semantic fidelity.
For each sample, participants independently conduct a pairwise comparison between the generated results of our method and baseline approaches, assigning a Good/Same/Bad (GSB) judgment, and the final label is determined by majority vote.
We report the ratio (G+S)/(B+S) as our evaluation metric, where higher scores indicate a stronger user preference.
This metric reflects the extent to which our method is judged as ``better or not worse'' than the baselines.
Beyond the \textbf{Overall Performance}, we conduct granular GSB assessments across five specific dimensions:

\begin{itemize}
    \item \textbf{Visual Quality.} 
    Evaluates the per-frame aesthetic quality, focusing on image sharpness, structural integrity, and the absence of generation artifacts within individual frames.
    
    \item \textbf{Dynamic Quality.} 
    Evaluates the temporal consistency of the generated video sequences, specifically examining motion smoothness, inter-frame coherence, and the stability of background and character elements across consecutive frames.
    
    \item \textbf{Identity Preservation.} 
    Measures how well the generated video maintains the recognizable identity traits and appearance details of the reference image throughout the animation.
    
    \item \textbf{Motion Accuracy.} 
    Examines the precision of overall body motion transfer, determining whether the generated poses and gestures accurately replicate the trajectory and amplitude of the driving video.
    
    \item \textbf{Expression Accuracy.} 
    Evaluates the alignment of facial dynamics with the driving source, including the accuracy of global head pose and the subtlety of micro-expressions.
\end{itemize}

This comprehensive GSB protocol provides a unified framework for evaluating critical aspects ranging from fine-grained visual details to holistic motion transfer, offering a reliable indicator of real-world user experience. We will also incorporate additional objective metrics in the future to complement and extend our quantitative evaluation.

\subsection{Experimental Results}

\begin{figure}[h]
    \centering
    \includegraphics[width=\linewidth]{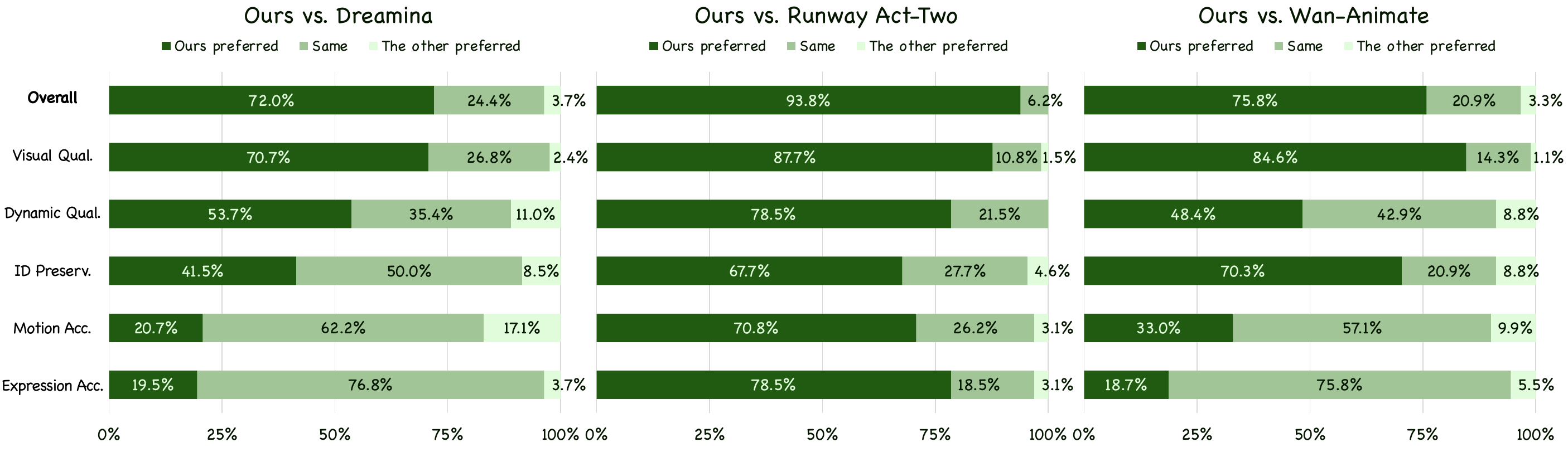}
    \caption{\textbf{Visualization of GSB evaluation results} (preference rates in percentages) comparing \methodname{} with Dreamina, Runway Act-Two, and Wan-Animate across various evaluation dimensions. Note that numerical labels are omitted for categories with 0\%.}
    \label{fig:gsb1}
\end{figure}

\myparagraph{Comparison with Baselines.}
For comparative evaluation, we select \textbf{Dreamina}~\citep{dreamina} and \textbf{Runway Act-Two}~\citep{runway_act_two} to represent the most competitive commercial solutions currently available on the market for holistic character animation.
Additionally, we include \textbf{Wan-Animate}~\citep{cheng2025wananimate}, which stands as the current state-of-the-art method among open-source approaches.
All methods are evaluated under a unified setting at 1080P resolution with the same video duration, strictly following the officially recommended best-practice inference configurations for each method. Text prompts are semantically aligned across all methods.
The GSB evaluation results against these competing baselines are summarized in Tab.~\ref{tab:comparison_gsb}, and the detailed distribution of GSB ratings is visualized in Fig.~\ref{fig:gsb1}.
Numerical results demonstrate that \methodname{} \textbf{surpasses all competitors across every evaluation dimension}, delivering superior or comparable performance.
In particular, \methodname{} achieves significantly higher scores in Overall Preference and Visual Quality compared to other approaches, demonstrating our superior capability in generating high-fidelity and visually coherent animated videos.

\begin{table}[h]
    \centering
    \caption{\textbf{Numerical results of GSB metrics} between \methodname{} and competitors across diverse criteria. ``Qual.'' is short for ``Quality''; ``Preserv'' is short for ``Preservation''; ``Acc.'' is short for ``Accuracy''.} 
    \resizebox{\linewidth}{!}{
        \begin{tabular}{@{}l|cccccc@{}}
        \toprule
        GSB & Overall & Visual Qual. & Dynamic Qual. & ID Preserv. & Motion Acc. & Expression Acc. \\ \midrule
        Ours vs. Dreamina & 3.44 & 3.33 & 1.92 & 1.56 & 1.05 & 1.20 \\
        Ours vs. Runway Act-Two & 16.25 & 8.00 & 4.64 & 2.95 & 3.32 & 4.50 \\
        Ours vs. Wan-Animate & 4.00 & 6.43 & 1.77 & 3.07 & 1.34 & 1.16 \\ \bottomrule
        \end{tabular}
    }
    \label{tab:comparison_gsb}
\end{table}

\begin{figure}[tp!]
    \centering
    \includegraphics[width=\linewidth]{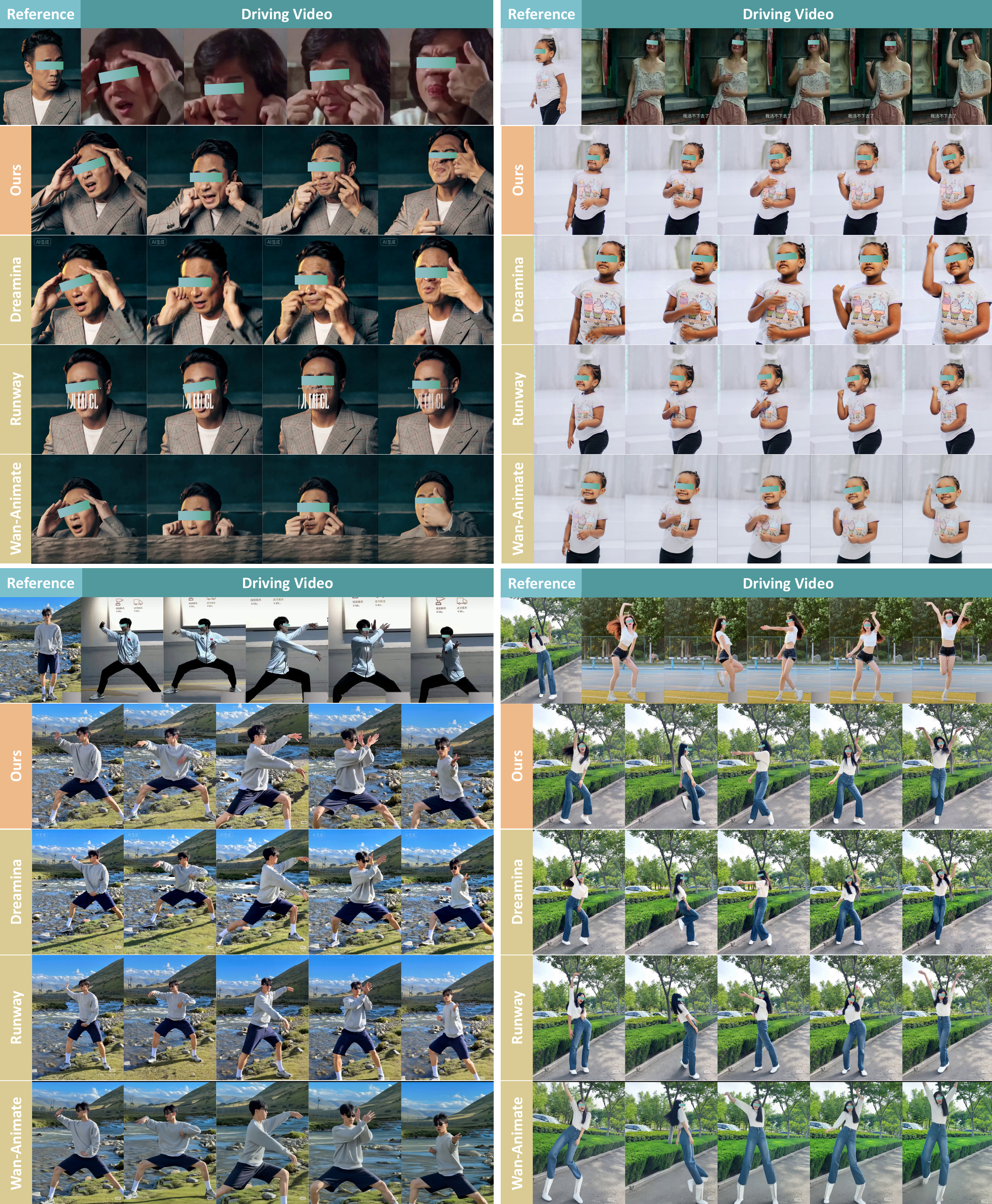}
    \caption{\textbf{Qualitative comparisons with baseline methods.}
    \methodname{} delivers high-fidelity holistic character animation videos, characterized by exceptional expressiveness and motion accuracy, while maintaining faithful identity and scene consistency.
    \textbf{Top:} Our method produces more vivid and precise facial expressions and hand gestures.
    \textbf{Bottom:} Our framework exhibits superior robustness against complex and rapid body motion, yielding higher-fidelity results.}
    \label{fig:comp}
\end{figure}

Qualitative comparisons are presented in Fig.~\ref{fig:comp}.
The top two cases highlight the superiority of our method in modeling and reproducing fine-grained facial expressions and hand gestures.
Specifically, Dreamina demonstrates limited expressiveness under extreme emotional states (e.g., intense sadness) and, similar to Wan-Animate, struggles with complex gestures, frequently resulting in erroneous hand movements and artifacts.
Runway Act-Two exhibits poor robustness against intricate hand poses and facial dynamics, even suffering a complete failure in the top-left case.
In contrast, \methodname{} achieves precise replication of both extreme and subtle expressions as well as complex hand interactions, yielding high-fidelity and expressive results.
Furthermore, competing methods often exhibit identity inconsistencies during cross-identity transfer.
For instance, in the top-left case, Dreamina alters the limb proportions of the child character.
Conversely, thanks to our specialized motion retargeting and identity disentanglement strategies, our method effectively handles significant discrepancies between the driving and reference subjects while faithfully preserving the reference identity.
The bottom two cases illustrate comparative performance under complex, large-amplitude, and rapid motion scenarios.
Specifically, Dreamina suffers from spatial depth ambiguity regarding limb relationships and exhibits artifacts characterized by incomplete or broken limb structures.
Runway is prone to catastrophic failure when confronting such challenging motion modeling tasks.
Meanwhile, Wan-Animate completely fails to reproduce rapid and challenging dynamics, accompanied by severe appearance degradation and significant global color drift.
In contrast, our method effectively handles these extreme conditions, generating precise and physically plausible motions without suffering from structural distortion or appearance drift.
In summary, these qualitative comparisons validate that \methodname{} excels in generating high-quality holistic body motions, successfully achieving natural and adaptive cross-identity transfer with highly preserved identity appearance and reference visual cues.

\begin{figure}[!p]
    \centering
    \includegraphics[width=\linewidth]{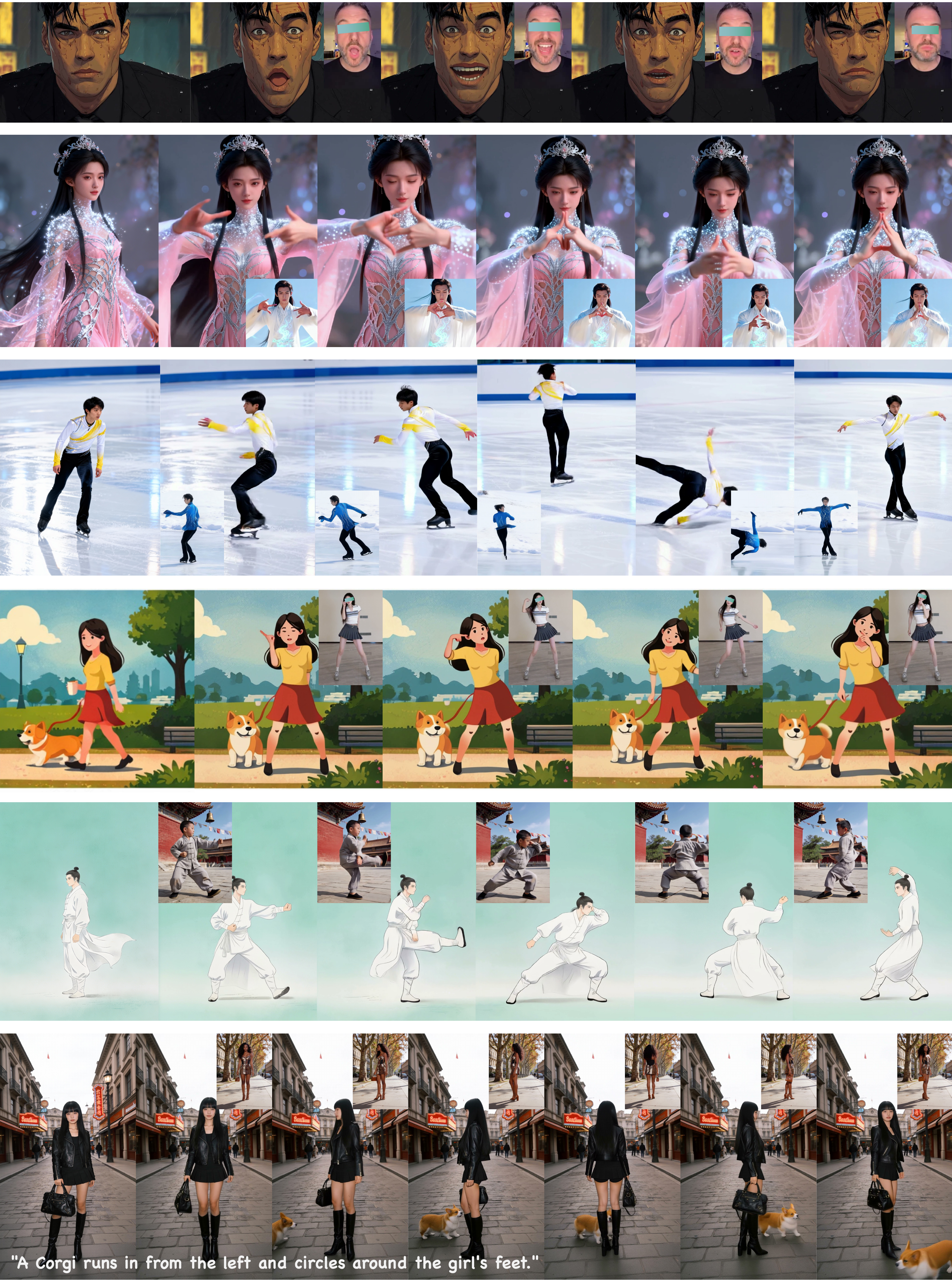}
    \caption{\textbf{Visualization of generated results across diverse scenarios.}
    We highlight \methodname{}'s capability to generate high-fidelity character animations with accurate motion imitation ranging from complex body dynamics to fine-grained facial and gestural details, while faithfully preserving appearance across various identities and maintaining excellent text controllability.
    The leftmost column displays reference images, with generated results and driving sequences (insets) in the remaining columns.}
    \label{fig:case}
\end{figure}

\myparagraph{Visualization results on diverse scenarios.}
Fig.~\ref{fig:case} further presents additional visualization results generated by \methodname{} across a diverse range of scenarios.
Benefiting from our unified framework that incorporates heterogeneous modeling tailored to the inherent characteristics of distinct body regions, \methodname{} successfully handles challenging large-scale full-body movements while simultaneously capturing fine-grained facial expressions, lip motions, and complex hand interactions across diverse shot scales, ranging from close-up portraits to full-body views.
Attributable to our meticulously designed adaptive motion retargeting and identity preservation implementations, our method facilitates natural, seamless, and robust motion transfer even between subjects with significant discrepancies in body shape and appearance (e.g., child-to-adult, realistic-to-cartoon).
Crucially, it strictly preserves the identity traits of the reference image without suffering from body distortions or other appearance artifacts.
Furthermore, leveraging our semantic motion guidance and prompt enhancer, \methodname{} maintains faithful text responsiveness alongside precise motion control and high-quality video generation.
This capability allows for flexible manipulation of attributes beyond the reference image—such as character clothing, background elements, and environmental changes—thereby significantly enhancing overall controllability.

\section{Related Work}
\subsection{Video Diffusion Models}
The advent of Diffusion Models~\citep{ho2020denoising} has revolutionized the landscape of video synthesis, enabling the generation of high-fidelity content with unprecedented controllability.
In particular, these models have been widely applied to human-centric video generation tasks~\citep{ding2025klingavatar,team2025klingavatar,he2024co,chen2025humo}, serving as a foundational technique for animate characters.
Early explorations in video diffusion models primarily extended pretrained image-based U-Net~\cite{dhariwal2021diffusion,rombach2022high} architectures by incorporating temporal modules such as 3D convolutions or temporal attention mechanisms to model cross-frame dependencies~\citep{blattmann2023stable,guo2023animatediff}.
Despite their initial success, these U-Net-based paradigms often face inherent limitations regarding scalability in both resolution and sequence duration.
Consequently, recent research has shifted towards Diffusion Transformers (DiTs)~\citep{peebles2023scalable} as the mainstream backbone.
By compressing videos into spatiotemporal tokens via 3D VAEs and leveraging the scalable attention mechanisms of Transformers, DiTs effectively capture long-range temporal dynamics and support stable large-scale generation\citep{kong2024hunyuanvideo,wan2025wan,kling2024,li2024sora}.
In this work, we build our holistic animation framework upon a robust DiT backbone, harnessing its powerful generative capabilities and rich internal priors regarding human structure and motion dynamics to facilitate precise character animation.

\subsection{Character Animation}
\myparagraph{Body animation.}
Pioneering efforts in body animation, such as FOMM~\citep{siarohin2019first} and MRAA~\citep{siarohin2021motion}, primarily relied on unsupervised optical flow estimation to warp source features driven by motion trajectories.
With the advent of Latent Diffusion Models (LDMs)~\citep{rombach2022high}, the field has shifted toward controllable synthesis using structural guidance.
A prominent stream utilizes explicit 2D skeletal poses for control; for instance, Animate Anyone~\citep{animateanyone} introduced a lightweight Pose Guider to encode skeleton signals, while MimicMotion~\citep{zhang2024mimicmotion} and Animate-X~\citep{tan2024animate} further improved robustness via confidence-aware guidance and explicit pose augmentation strategies.
Alternative approaches seek to incorporate 3D priors or surface details: MagicAnimate~\citep{xu2024magicanimate} employs DensePose~\citep{guler2018densepose} to establish dense correspondences, whereas Champ~\citep{zhu2024champ} and MagicMan~\citep{he2025magicman} leverage 3D parametric models (SMPL)~\citep{SMPL:2015} to enforce geometric consistency.
More recently, research has expanded to handle complex human-scene interactions~\citep{men2025mimo,hu2025animate} and is transitioning backbone architectures from U-Nets to Diffusion Transformers (DiTs)~\citep{cheng2025wananimate,wang2025unianimatedit}.
In this work, we design a tailored and robust representation to model global body dynamics with high accuracy and expressiveness.
Crucially, we further complement this body modeling by refining hand regions with a representation adapted to hand characteristics, effectively reconciling global stability with fine-grained articulatory details.

\myparagraph{Facial animation.}
Early approaches primarily leveraged GANs, utilizing neural keypoints or 3D parametric models to drive portrait animation via image warping.
Notably, LivePortrait~\citep{guo2024liveportrait} recently achieved impressive real-time performance and control.
Recently, the paradigm shift to Diffusion Models has significantly elevated generation quality.
XPortrait~\citep{xie2024xportrait} enhanced cross-identity reenactment through patch-based local control, while SkyReels-A1~\citep{qiu2025skyreels} demonstrated the scalability of Diffusion Transformers (DiTs) for high-resolution portrait synthesis.
X-Nemo~\citep{zhao2025x} further introduced 1D latent descriptors to represent appearance-agnostic facial dynamics.
Distinct from these methods, we carefully design an adaptive representation to model the rich and unstructured facial dynamics.
This modeling is further optimized to effectively filter out unnecessary identity cues while sharpening the capture of critical motion patterns, ultimately achieving superior identity disentanglement and micro-expression fidelity.

\myparagraph{Holistic full-body animation.}
Recent pioneering works, including Wan-Animate~\citep{cheng2025wananimate}, X-UniMotion~\citep{song2025xuni}, and DreamActor-M1~\citep{luo2025dreamactor}, have initiated the exploration of holistic full-body animation.
However, these methods still struggle to effectively coordinate motions across varying granularities and achieve robust identity-motion disentanglement and retargeting within a unified framework.
Consequently, they often suffer from visual artifacts and severe identity drift, particularly when confronting challenging articulations or conducting cross-identity motion transfer.
Addressing these critical limitations constitutes the primary focus of our work.

\footnotetext{Individuals and characters shown in this paper are for scientific visualization and technical illustration purposes only.}

\section{Conclusion}
In this paper, we present \textbf{\methodname{}}, a unified DiT-based framework that achieves robust, precise, and expressive holistic character animation.
By orchestrating heterogeneous motion representations tailored to the distinct characteristics of body, facial, and hand dynamics, our approach successfully achieves both large-scale structural stability and fine-grained articulatory expressiveness within a single cohesive system.
Furthermore, we address the challenge of cross-identity transfer through adaptive motion learning, enabling natural adaptation to diverse characters.
Through meticulously designed identity representation encoding and fusion, \methodname{} effectively maintains subject identity consistency during motion transfer, and an innovative subject library allows users to provide additional reference identity information to further enhance appearance fidelity.
Additionally, our advanced acceleration strategies boost inference efficiency by over 10$\times$, which, coupled with intelligent prompt enhancement, significantly elevates practical utility and multi-conditional controllability. Qualitative results and numerical human preference evaluations demonstrate that \methodname{} outperforms state-of-the-art commercial and open-source solutions, delivering efficient and high-fidelity animation results characterized by precise holistic control, exceptional robustness against rapid dynamics, and faithful appearance preservation.

\section*{Impact Statement}
The rapid evolution of character animation presented in this work offers significant potential for transforming digital entertainment, virtual reality, and creative content production. However, the capability to synthesize highly realistic human videos also brings forth critical ethical considerations. The precise control over body dynamics and facial reenactment, combined with faithful identity preservation, raises concerns regarding the potential for privacy violations, the unauthorized appropriation of likeness, and the risk of creating deceptive ``deepfake'' media.

As with advanced generative technologies, there is a possibility that these methods could be misused to animate individuals without their consent, synthesizing actions or statements they never performed. Addressing these risks requires not only technical safeguards but also the collective development of ethical guidelines and legal frameworks. In this work, we are steadfastly committed to responsible research practices. We advocate for the implementation of safety mechanisms, such as content filtering and watermarking, to prevent misuse. All data processing and model development described herein adhere to strict ethical standards, intended to advance the fields of computer vision and graphics. We believe that by upholding these principles, we can harness the creative power of motion control technology while safeguarding societal trust and individual rights.

\section*{Contributors}
All contributors are listed in alphabetical order by their last names.

Jialu Chen, Yikang Ding, Zhixue Fang, Kun Gai, Kang He, Xu He, Jingyun Hua, Mingming Lao, Xiaohan Li, Hui Liu, Jiwen Liu, Xiaoqiang Liu, Fan Shi, Xiaoyu Shi, Peiqin Sun, Songlin Tang, Pengfei Wan, Tiancheng Wen, Zhiyong Wu, Haoxian Zhang\footnote{Project Lead}, Runze Zhao, Yuanxing Zhang, Yan Zhou

\clearpage
\newpage
\bibliographystyle{kling/plainnat}
\bibliography{main}

\clearpage
\newpage

\end{document}